\documentclass[12pt,a4paper,oneside]{article} 

\usepackage[english]{babel} 
\usepackage[T1]{fontenc} 
\usepackage[utf8]{inputenc} 
\usepackage[babel]{microtype} 

\usepackage{amsmath,amssymb,bm,mathtools}
\usepackage{enumitem}
\usepackage{booktabs}
\usepackage[detect-weight=true,detect-family=true]{siunitx}
\usepackage{url}
\usepackage[pdftex]{graphicx}
\graphicspath{{fig/}}
\usepackage{hyperref}
\usepackage{fullpage}
\usepackage{caption} 
\newlist{enuminline}{enumerate*}{1}
\setlist[enuminline,1]{label=\itshape\alph*\upshape)}

\newcommand{\email}[1]{\href{mailto:#1}{\nolinkurl{#1}}}

\title{The Player Kernel: Learning Team Strengths\\
Based on Implicit Player Contributions}
\author{
Lucas Maystre
\and
Victor Kristof
\and
Antonio J. González Ferrer
\and
Matthias Grossglauser\\
\email{firstname.lastname@epfl.ch}
}

\date{}

\begin{document}
\maketitle

\begin{abstract}
In this work, we draw attention to a connection between skill-based models of game outcomes and Gaussian process classification models.
The Gaussian process perspective enables
\begin{enuminline}
\item a principled way of dealing with uncertainty and
\item rich models, specified through kernel functions.
\end{enuminline}
Using this connection, we tackle the problem of predicting outcomes of football matches between national teams.
We develop a \emph{player kernel} that relates any two football matches through the players lined up on the field.
This makes it possible to share knowledge gained from observing matches between clubs (available in large quantities) and matches between national teams (available only in limited quantities).
We evaluate our approach on the Euro 2008, 2012 and 2016 final tournaments.
\end{abstract}

\section{Introduction}
\label{sec:intro}


Statistical models of game outcomes have a rich and diverse history, going back almost a century:
as early as 1928, Zermelo \cite{zermelo1928berechnung} proposed a simple algorithm that infers the skill of chess players based on observed game outcomes.
Zermelo's ideas have since been rediscovered and refined multiple times, and have been successfully applied to various sports-related prediction problems and beyond.
On the occasion of the Euro 2016 football tournament, we revisit these ideas and highlight their connections to modern machine learning techniques.
In particular, we show how Zermelo's model can be cast as a Gaussian process classification model.
The Gaussian process framework provides two key advantages.
First, it brings all the benefits of Bayesian inference. In particular it provides a principled way to deal with the uncertainty associated to noisy observations and to predictions.
Second, it opens up new modeling perspectives through the specification of kernel functions.

\begin{figure}[t]
  \centering
  \includegraphics[width=\linewidth]{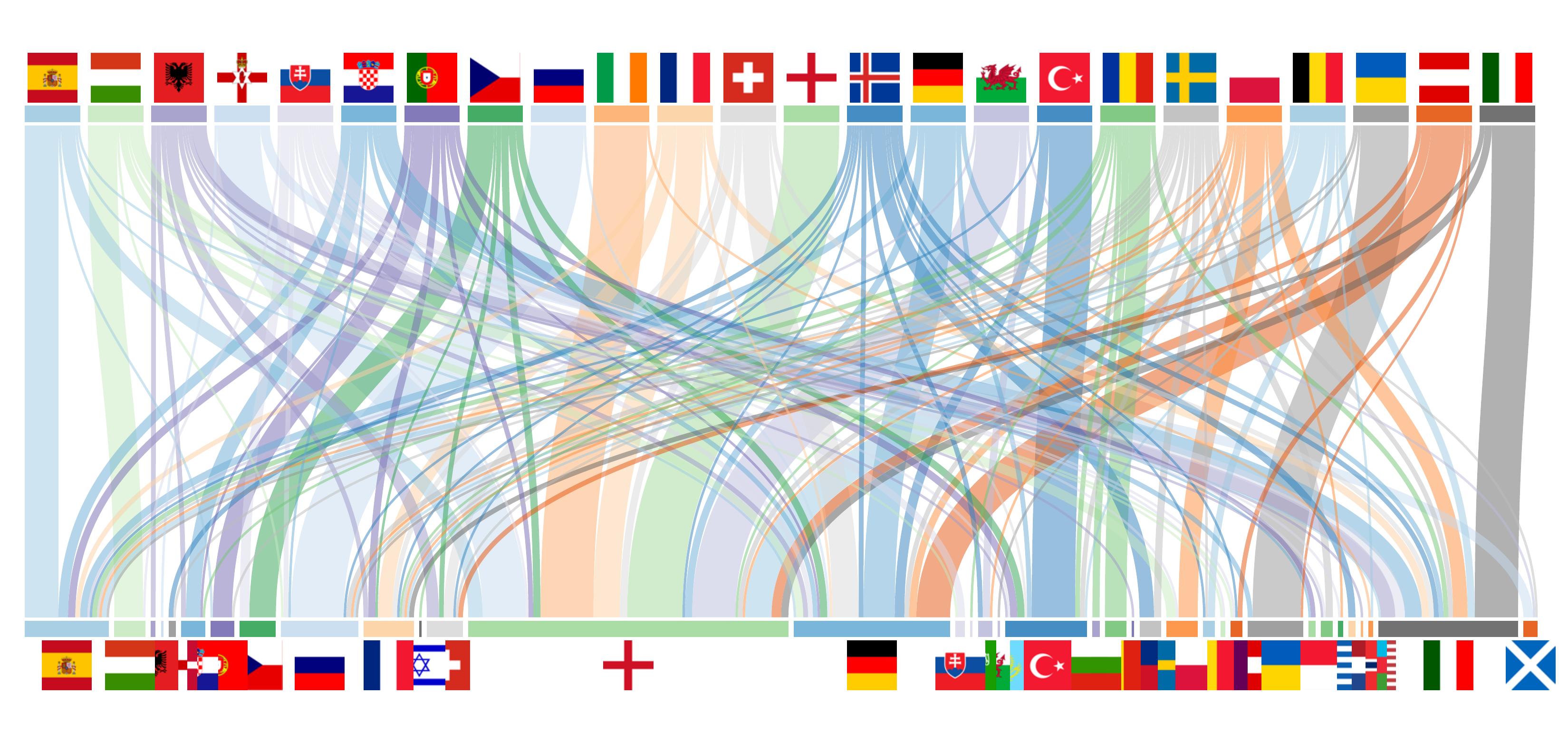}
  \caption{
  Players of national teams qualified for the Euro 2016 (top row) are playing in clubs across Europe and beyond (bottom row).
  The English, German and Italian club championships contain the most selected players.
}
  \label{fig:sankey}
  \vspace{-0.1cm}
\end{figure}

Equipped with this, we investigate the problem of predicting outcomes of football matches between national teams.
We identify two key challenges,
\begin{enuminline}
\item that of \emph{data sparsity} (national teams usually play no more than ten matches per year), and
\item that of \emph{data staleness} (the team roster is constantly evolving).
\end{enuminline}
Taking inspiration from the observation that national teams' players frequently face each other in competitions between clubs (see Figure~\ref{fig:sankey}), we show that these two difficulties can be tackled by the introduction of a \emph{player kernel}.
This kernel relates any two matches through the players lined up on the field, and makes it possible to seamlessly use matches between clubs to improve a predictive model ultimately used for matches between national teams.
In contrast to national teams, clubs play much more frequently, and more data is available.

The remainder of this short report is organized as follows.
We review related work at the end of the present section.
In Section~\ref{sec:methods}, we formalize the link between Zermelo's ideas and Gaussian processes, and present our player kernel.
Then, in Section~\ref{sec:evaluation}, we evaluate our predictive model on the Euro 2008, 2012 and 2016 final tournaments.

\subsection{Related Work}

More than two decades after Zermelo's seminal paper \cite{zermelo1928berechnung}, his model for paired comparisons was rediscovered and popularized by Bradley and Terry \cite{bradley1952rank}.
Nowadays, the model is usually referred to as the Bradley--Terry model.
In the context of skill-based game modeling, the same model (associated to a simple online stochastic gradient update rule) is also known as the Elo rating system \cite{elo1978rating}.
It is used by FIDE to rank chess players\footnote{See: \url{https://ratings.fide.com/}.} and by FIFA to rank women national football teams\footnote{See: \url{http://www.fifa.com/fifa-world-ranking/procedure/women.html}.}, among others.

The model and related inference algorithms have been extended in various ways; one direction that is of particular interest is the handling of uncertainty of the estimated skill parameters.
Glickman \cite{glickman1999parameter} proposes an extension that simultaneously updates ratings and associated uncertainty values after each observation.
Herbrich et al. \cite{herbrich2006trueskill} propose TrueSkill, a comprehensive Bayesian framework for estimating player skill in various types of games.
The models and methods described in this paper are fundamentally similar to TrueSkill, as will be discussed in Section~\ref{sec:methods}.
Finally, in the context of learning users' preferences from pairwise comparisons, Chu and Ghahramani \cite{chu2005preference} present a Gaussian process approach that is comparable to our work.

\section{Methods}
\label{sec:methods}

In this section, we first show how the model of pairwise comparisons proposed by Zermelo \cite{zermelo1928berechnung} and popularized by Bradley and Terry \cite{bradley1952rank} and Elo \cite{elo1978rating} can be expressed in the Gaussian process framework.
Second, we present the player kernel, a covariance function that relates matches through lineups.

\subsection{Pairwise Comparisons as Gaussian Process Classification}

Suppose that we observe outcomes of comparisons between two objects (e.g., two players or two teams) in a universe of objects denoted $1, \ldots, M$.
We begin by restricting ourselves to binary outcomes, i.e., we assume that one of the two objects wins.
Zermelo \cite{zermelo1928berechnung} postulates that each object $u$ can be represented by a parameter $w_u \in \mathbf{R}_{>0}$, indicative of its relative chances of winning against an opponent.
Given these parameters, the probability of observing the outcome ``$u$ wins against $v$'' (denoted by $u \succ v$) is given by $w_u / (w_u + w_v)$.
Using the reparametrization $w_u = e^{s_u}$, this can be rewritten as
\begin{align}
\label{eq:logistic}
P(u \succ v) = \frac{1}{1 + \exp[-(s_u - s_v)]} = \frac{1}{1 + \exp(- \bm{s}^\top \bm{x})},
\end{align}
where $\bm{s} = [s_i]$ and $\bm{x} \in \mathbf{R}^M$ is such that $x_u = 1$, $x_v = -1$ and $x_i = 0$ for $i \ne u, v$.
As such, the pairwise comparison model can be seen as a special case of logistic regression, where the feature vector simply indicates the winning and losing objects.
Furthermore, logistic regression is itself a special case of Gaussian process classification \cite[Ch. 3]{rasmussen2006gaussian}.
A Gaussian process $f(\bm{x}) \sim \mathcal{GP}(m(\bm{x}), k(\bm{x}, \bm{x}'))$ is defined by a mean function $m(\bm{x})$ and a positive semi-definite covariance (or kernel) function $k(\bm{x}, \bm{x}')$.
Given any finite collection of points $\bm{x}_1, \ldots, \bm{x}_N$, the Gaussian process sampled at these points has a multivariate Gaussian distribution
\begin{align*}
\begin{bmatrix}
f(\bm{x}_1) & \dots & f(\bm{x}_k)
\end{bmatrix} = \mathcal{N}(\bm{m}, \bm{K}),
\end{align*}
where $m_i = m(x_i)$ and $K_{ij} = k(\bm{x}_i, \bm{x}_j)$.
It is not hard to show that if $\bm{s} \sim \mathcal{N}(\bm{0}, \sigma^2 \bm{I})$, then $f(\bm{x}) = \bm{s}^\top \bm{x}$ is a Gaussian process with $m(\bm{x}) = 0$ and $k(\bm{x}, \bm{x}') = \sigma^2 \bm{x}^\top \bm{x}'$.
This enables the interpretation of \eqref{eq:logistic} as the likelihood of a Gaussian process classification model with the logit link function.

The Gaussian process viewpoint shifts the focus from the representation of the function $f(\bm{x})$ (in the case of \eqref{eq:logistic}, a linear function) to the correlation between two function evaluations, as defined by the kernel function $k(\bm{x}, \bm{x}')$.
Intuitively, the model can simply be specified by how similar any two match outcomes are expected to be.
Furthermore, the Gaussian process viewpoint also makes it possible to take advantage of the vast amount of literature and software related to accurate, efficient and scalable inference.

\paragraph{Handling draws.}
Rao and Kupper \cite{rao1967ties} propose an extension of the pairwise comparison model for ternary (win, draw, loss) outcomes.
In this extension, the two different types of outcomes have probabilities
\begin{align*}
P(u \succ v) = \frac{1}{1 + \exp[f(\bm{x}) - \alpha]}\ \text{and}\ 
P(u \equiv v) = (e^{2 \alpha} - 1) P(u \succ v) P(v \succ u),
\end{align*}
where $\alpha > 0$ is an additional hyperparameter controlling the draws.
Because a draw can be written as the product of a win and a loss, model inference can still be performed using only a \emph{binary} Gaussian process classification model, with minimal changes needed to the link function.

\subsection{The Player Kernel}

\newcommand{\Wx}{\ensuremath{\mathcal{W}}}
\newcommand{\Lx}{\ensuremath{\mathcal{L}}}
We now consider an application to football and propose a method to quantify how similar two match outcomes are expected to be.
Denote by $P$ the number of distinct players appearing in a dataset of matches.
We define a team's \emph{lineup} as the set consisting of the \num{11} players starting the match.
For a given match, let $\Wx$ and $\Lx$ be the lineups of the winning and losing teams, respectively.
Define $\bm{z} \in \mathbf{R}^P$ such that $z_p = 1$ if $p \in \Wx$, $z_p = -1$ if $p \in \Lx$ and $z_p = 0$ otherwise.
We then define the player kernel as
\begin{align*}
k(\bm{z}, \bm{z}') = \sigma^2 \bm{z}^\top \bm{z}'.
\end{align*}
Intuitively, the function is positive if the same players are lined up in both matches, and the same players win (respectively lose).
The function is negative when players win one match, but lose the other.
Finally, the function is zero, e.g., when the lineups are completely disjoint.

\begin{figure}[t]
  \centering
  \includegraphics[width=.8\linewidth]{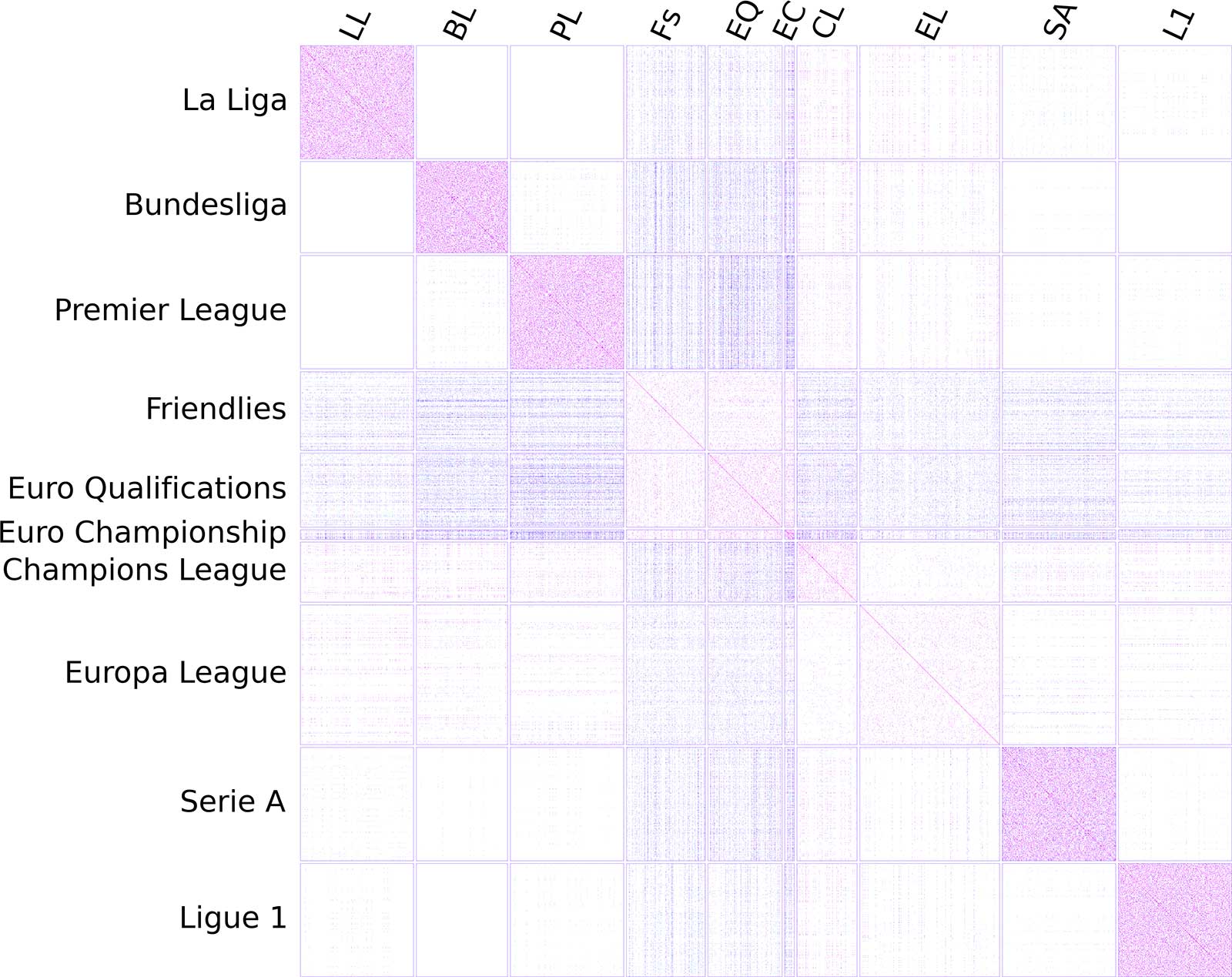}
  \caption{Heatmap of the magnitude of the kernel matrix for \num{3184} matches played over the year preceding Euro 2012.
White indicates zero correlation, a saturated color indicates more correlation.
Matches between national teams exhibit non-zero covariance with matches of all other competitions.
}
  \label{fig:kernel}
  \vspace{-0.1cm}
\end{figure}

This kernel implicitly projects every match into the space of players, and defines a notion of similarity in this space.
In the case of national teams qualified to Euro final tournaments, we find that this approach is very useful: a significant part of national teams' players take part in one of the main European leagues and play with or against each other.
International club competitions (such as the UEFA Champions League) further contribute to the ``connectivity'' among players.
Figure~\ref{fig:kernel} illustrates the similarity of matches across different competitions in 2011--2012.

It is interesting to note that the player kernel corresponds to a linear model over the players.
That is, it is equivalent to assuming that there is one independent skill parameter per player, and that the strength of a team is the sum of its players' skills.
Such a model contains a massive number of parameters (possibly much more than the number of observations), and there is little hope to reliably estimate every parameter.
In fact, we observe that the model is ``weakly'' parametric: the number of distinct players usually grows with the number of matches observed.
The kernel-based viewpoint that we take emphasizes the fact that estimating these parameters is not necessary.

\paragraph{Relation to TrueSkill.}
Our Gaussian process model coupled to the player kernel is very similar to TrueSkill \cite{herbrich2006trueskill}.
The most important difference is that we take advantage of the dual representation and operate in the space of matches instead of the space of players.
Beyond the conceptual reasons outlined above, it makes inference significantly less computationally intensive for the datasets that we consider.

\section{Experimental Evaluation}
\label{sec:evaluation}

In this section, we evaluate our predictive model on the matches of the Euro 2008, 2012 and 2016 final tournaments and compare it to several baselines.

We collect a dataset of matches from
\begin{enuminline}
\item official and friendly competitions involving national teams, and
\item the most prestigious European club competitions,
\end{enuminline}
starting from July 1\textsuperscript{st}, 2006.
There are approximately $15 \times$ more matches between clubs than there are matches between national teams in our dataset.
With respect to the model outlined in Section~\ref{sec:methods}, our final predictive model processes one additional feature that encodes which team played at home (this feature is null for matches played on neutral ground).
We train the model using all $N$ matches that were played prior to the start of the competition on which we test.
When computing the kernel matrix (whether on training or on test data) we use the starting lineups, usually announced shortly before the start of the game.
It is interesting to note that the number of distinct players $P$ appearing in the dataset exceeds the number of training instances in each case (the values of $N$ and $P$ are shown in Table~\ref{tab:eval}).
We use the GPy Python library\footnote{See: \url{https://sheffieldml.github.io/GPy/}.} to fit the model; inference takes a minute for the 2008 test set (17 minutes for 2016).
The predictions come in the form of probability distributions $[p^{\text{W}}, p^{\text{D}}, p^{\text{L}}]$ over the three outcomes (win, draw, loss).

We compare our predictive distributions against three baselines.
First, we consider a simple Rao-Kupper model based on national team ratings obtained from a popular Web site\footnote{See: \url{http://www.eloratings.net/}.}.
This model is similar to ours, but
\begin{enuminline}
\item it does not relate matches through player, and thus does not consider club outcomes, and
\item as ratings are fixed values, it does not consider uncertainty in the ratings.
\end{enuminline}
Second, we consider average probabilities derived from the odds given by three large betting companies.
Third, we consider a random baseline which always outputs $[1/3, 1/3, 1/3]$.
The predictive distributions are evaluated using the average logarithmic loss over $T$ test instances
\begin{align*}
- \frac{1}{T} \sum_{i=1}^{T} \left[
\mathbf{1}_{\{y_i = \text{W}\}} \log p^{\text{W}}_i
+ \mathbf{1}_{\{y_i = \text{D}\}} \log p^{\text{D}}_i
+ \mathbf{1}_{\{y_i = \text{L}\}} \log p^{\text{L}}_i
\right].
\end{align*}
The logarithmic loss penalizes more strongly predictions that are both confident and incorrect.
Table~\ref{tab:eval} summarizes the results.

\begin{table}[t]
  \caption{
  Average logarithmic loss of our predictive model (PlayerKern), a model based on national team ratings (Elo), betting odds (Odds) and a random baseline (Random) on the final tournaments of three European championships.
  $N$ is the number of training instances, $P$ the number of distinct players and $T$ the number of test instances.}
  \label{tab:eval}
  \centering
  \setlength\tabcolsep{0.2cm}
  \begin{tabular}{l rr rrrrr}
    \toprule
    Competition & $N$         & $P$         & $T$      & PlayerKern           & Elo                  & Odds                 &  Random \\
    \midrule
    Euro 2008   & \num{4390}  & \num{7875}  & \num{31} & \num{0.969}          & \textbf{\num{0.910}} & \num{0.979}          & \num{1.099} \\
    Euro 2012   & \num{15594} & \num{21735} & \num{31} & \textbf{\num{0.939}} & \num{1.003}          & \num{0.953}          & \num{1.099} \\
    Euro 2016   & \num{24887} & \num{33157} & \num{51} & \num{1.067}          & \num{1.102}          & \textbf{\num{1.020}} & \num{1.099} \\
    \bottomrule
  \end{tabular}
\end{table}

Our predictive model performs well in 2008 and 2012, but slightly less so in 2016.
It is noteworthy that the 2016 final tournament has been generally less predictable than earlier editions.
The case of the Elo baseline is interesting, as its accuracy varies wildly.
Reasons for this might include the noise due to the online gradient updates, and the lack of proper uncertainty quantification in the ratings.
Our method, in contrast, seems to produce more conservative predictions, but manages to achieve a more consistent performance

\subsection{Conclusion}

In this short report, we exposed a connection between a well-known pairwise comparison model and Gaussian process classification, and proposed a kernel that is able to transfer knowledge across different types of football matches---those between clubs and those between national teams.
We showed that a predictive model building on these ideas achieves a logarithmic loss that is competitive with betting odds.
In future work, we would like to investigate how to incorporate aging into the model, i.e., how to progressively downweight older data.

\subsubsection*{Acknowledgments}

The authors thank Young-Jun Ko for useful discussions at the early stages of the project.
\pagebreak

\hyphenation{wahrscheinlichkeits-rechnung} 
\bibliography{playerkern}

\end{document}